\documentclass{article}
\usepackage{ijcai26}

\usepackage{times}
\usepackage{soul}
\usepackage{url}
\usepackage[hidelinks]{hyperref}
\usepackage[utf8]{inputenc}
\usepackage[small]{caption}
\usepackage{graphicx}
\usepackage{amsmath}
\usepackage{amsthm}
\usepackage{amssymb}
\usepackage{booktabs}
\usepackage{algorithm}
\usepackage{algorithmic}
\usepackage{multirow}
\usepackage{subcaption}
\usepackage[switch]{lineno}

\title{Personalized Digital Health Modeling with Adaptive Support Users}

\author{
Zhongqi Yang$^1$
\and
Mahkameh Rasouli$^3$\and
Neda Mohseni$^1$\and
Yong Huang$^1$\and
Iman Azimi$^2$\And \\
Amir M. Rahmani$^{1,3}$\\
\affiliations
$^1$Department of Computer Science, University of California, Irvine\\
$^2$Thrive AI Health\\
$^3$Sue \& Bill Gross School of Nursing
, University of California, Irvine\\\
\emails
\{zhongqy4,mahkamer,nmohsen1,yongh7\}@uci.edu,
iman@thriveaihealth.com,
a.rahmani@uci.edu
}

\begin{document}

\maketitle

\begin{abstract}
Personalized models are essential in digital health because individuals exhibit substantial physiological and behavioral heterogeneity. 
Yet personalization is limited by scarce and noisy user-specific data. 
Most existing methods rely on population pretraining or data from similar users only, which can lead to biased transfer and weak generalization.
We propose a unified personalization framework that trains a personal model using adaptively weighted support users, including both similar and dissimilar individuals. 
The objective integrates personal loss, similarity-weighted transfer from similar users, and contrastive regularization from dissimilar users to suppress misleading correlations. 
An iterative optimization algorithm jointly updates model parameters and user similarity weights.
Experiments on six tasks across four real-world digital health datasets show consistent improvements over population and personalized baselines. 
The method achieves up to 10\% lower RMSE on large-scale datasets and approximately 25\% lower RMSE in low-data settings. 
The learned adaptive weights improve data efficiency and provide interpretable guidance for targeted data selection.
\end{abstract}

\section{Introduction}

Personalized machine learning is critical in digital health because physiological, behavioral, and intervention responses vary substantially across individuals~\cite{kolluri2016revolutionizing,johnson2021precision}. 
Models tailored to a single person can better capture these personal patterns than population-level predictors, enabling more accurate monitoring and recommendations~\cite{yang2024integrating,tazarv2021personalized,yang2024chatdiet,yang2025personalized}. 
However, effective personalization is difficult because personal data are often limited, noisy, and fragmented. 
When individual data are insufficient to train a stable model, external data become necessary~\cite{siirtola2019incremental}, yet incorporating them without distorting the personal signal remains a fundamental challenge.

Most existing approaches address data scarcity through population pretraining followed by fine-tuning, or by pooling data from users deemed similar to the target individual~\cite{wu2023personalized,yang2023loneliness,lee2023fedl2p,jain2021differentially}. 
These strategies rely almost exclusively on similar users and differ mainly in how similarity is enforced. 
Population pretraining treats all external data uniformly regardless of relevance, while similarity-based methods restrict transfer to a small neighborhood. 
In both cases, dissimilar users whose data may encode informative contrasts or counterexamples are excluded, leading to biased transfer and limited coverage of population-level variation.

This paper adopts a different perspective. We argue that personal models should be trained using the data of support users, which we define as a group that includes both similar and dissimilar individuals. 
Similar users provide aligned patterns that reinforce personal tendencies, while dissimilar users offer complementary information that helps the model understand what the target user is not. These contrasting profiles supply a rich counterfactual structure that purely similar-based approaches lack. Incorporating such diversity is particularly important in digital health because atypical physiology, marginal behavioral patterns, and rare fluctuations carry meaningful predictive signals. However, directly including data from dissimilar users may degrade accuracy by introducing misleading patterns and biases~\cite{hahn2022connecting,zhang2024gpfedrec}. 
There is a need for a training method that can adaptively leverage external data so that the model can transfer beneficial information and suppress detrimental signals in a principled manner.

Building on this insight, we introduce a method that adaptively identifies and integrates support users during personal model training. 
The framework quantifies the relevance of each external user to the target individual and modulates their influence through learned adaptive weights. 
It formulates personalization as a unified objective that combines a personal loss based on the target user’s own data, a similarity-weighted knowledge transfer term from similar users, and a contrastive regularization term from dissimilar users. Similar users contribute positive transfer that supports generalization, while dissimilar users provide structured contrast that regularizes the personal model and mitigates misleading correlations. 
To balance these components, the method learns adaptive weights that determine each support user’s contribution throughout training. 
We introduce an iterative optimization algorithm that alternates between updating model parameters and refining these adaptive weights. We evaluate the framework across six health monitoring tasks drawn from four real-world digital health datasets and compare its performance on two widely used deep learning architectures against representative personalization baselines.
We also examine data efficiency by relying only on support users and show that the learned adaptive weights offer interpretability and actionable insights for improving personal model development.

Our contributions are as follows:
\begin{enumerate}
\item We introduce the notion of support users comprising both similar and dissimilar individuals and explain why dissimilar users, although overlooked in prior work, provide meaningful complementary information for personal model training.
\item We propose a unified framework for integrating heterogeneous support user data. The framework learns how to leverage support users by assigning adaptive contributions from each external individual, which allows the model to benefit from helpful signals and suppress detrimental ones.
\item We demonstrate extensive empirical validation across six healthcare tasks using four real-world digital health datasets. We show that the adaptive weighting mechanism enhances personal model interpretability and provides actionable insights for targeted data selection.
\end{enumerate}

\section{Related Work}

Personalized machine learning has been extensively studied in federated learning and meta-learning. Existing approaches primarily differ in how they leverage data from other users during training and personalization. Below, we review three representative paradigms most closely related to this work.

\subsection{Global and Transfer-Based Personalization}
A common personalization strategy trains a single population-level model using data from all users and then adapts it to each individual through fine-tuning or regularization. 
This paradigm is exemplified by approaches that learn a shared global model and personalize it locally, either by updating all parameters or restricting personalization to specific layers \cite{mcmahan2017communication,oh2021fedbabu,li2021fedbn}. 
More advanced formulations introduce explicit coupling between global and personal objectives, encouraging local models to remain close to a shared reference while allowing individual adaptation \cite{li2021ditto,t2020personalized,deng2020adaptive}.
Meta-learning-based approaches similarly aim to learn a population-level initialization that enables rapid personalization using limited local data \cite{fallah2020personalized,chen2018federated}. Despite their effectiveness, these methods treat the population as a homogeneous source of knowledge and aggregate information uniformly across users, making them vulnerable to negative transfer when user distributions are highly heterogeneous.

\subsection{Clustered and Multi-Model Personalization}
To address cross-user heterogeneity, a second line of work partitions users into clusters or maintains multiple expert models, restricting parameter sharing to users with similar data distributions. Clustered federated learning methods alternate between assigning users to clusters and training a separate model per cluster, thereby isolating incompatible users \cite{liu2023auxo}. Related approaches employ mixture or ensemble formulations, where multiple experts are trained and users either select or weight these experts based on local validation performance \cite{yi2024pfedmoe,zhang2020personalized}. By limiting collaboration to similar users, these methods reduce negative transfer but also discard data from dissimilar users entirely. As a result, potentially informative complementary signals from atypical or contrasting users are not exploited once clustering or expert assignment stabilizes.

\subsection{Similarity-Weighted Personalization}
More recent work replaces hard clustering with soft, similarity-aware collaboration mechanisms that assign continuous weights to other users’ contributions. These methods typically compute pairwise similarity scores and use them to modulate information exchange, suppressing influence from distant users while emphasizing closer ones \cite{huang2021personalized}. Attention-based formulations further automate relevance estimation by learning interaction weights among users or models \cite{chen2024personalized}. A small number of studies have begun to question the assumption that dissimilar users are always harmful, suggesting that diversity among clients can sometimes improve personalization \cite{wu2024diversity}. However, existing similarity-weighted approaches primarily focus on attenuating the impact of dissimilar users rather than selectively leveraging them, and they rely on fixed or weakly adaptive weighting schemes without explicitly modeling the distinct roles of similar and dissimilar users.


%

\section{Adaptive Support-User Optimization Framework}

We propose a personalization framework that trains a personal model by adaptively leveraging data from support users, defined to include both similar and dissimilar individuals. 
Similar users provide aligned predictive structure that reinforces personal patterns, while dissimilar users supply structured contrast that helps suppress misleading population signals. 
The framework formulates personalization as a single optimization objective composed of three terms: a personal loss on the target user’s data, a similarity-weighted transfer loss from aligned users, and a contrastive regularization term from dissimilar users. 
Each support user’s contribution is modulated by learned adaptive weights, which are jointly optimized with model parameters.

\begin{figure*}[bt]
    \centering
    \includegraphics[width=\linewidth]{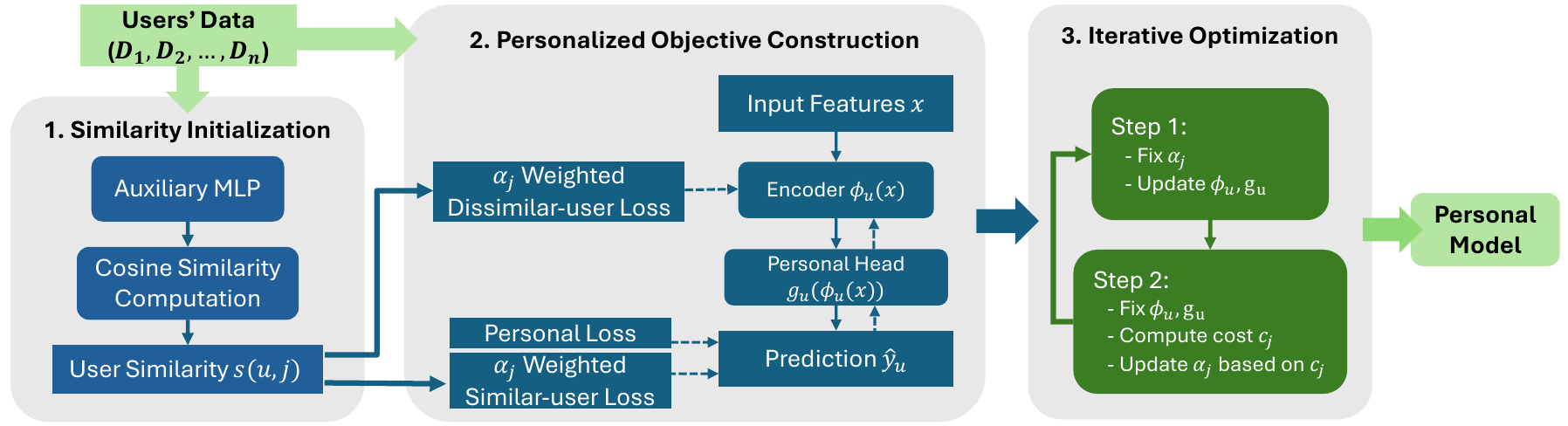}

\caption{
The personal model is built for each user through three stages: (1) user similarity initialization, (2) personalized loss construction with similar and dissimilar users, and (3) iterative optimization that alternates between model updates and adaptive weight refinement.
}

    \label{fig:pipeline}
    
\end{figure*}

Figure~\ref{fig:pipeline} provides an overview of the training procedure. 
The method first computes similarity scores between the target user and external users to initialize the support set. 
It then constructs the unified personalization objective and performs iterative optimization that alternates between updating model parameters and refining adaptive weights. 
The following subsections describe these components in detail.

\subsection{Support User Initialization through Similarity}

The first step initializes the set of support users by computing similarity scores between the target user and each external individual. These scores provide an initial estimate of which users are likely to contribute aligned signals and which users may provide contrasting information. For a target user \( u \) and another user \( j \), we compute a similarity value \( s(u,j) \in [0,1] \), where larger values indicate stronger similarity. All users with available data are retained as potential support users because both highly similar and highly dissimilar users will be incorporated through the adaptive mechanism.

In our experiments, we compute user similarity using cosine similarity between user embeddings extracted from an auxiliary neural network trained on pooled population data. Section \ref{similarity_compute} provides details on the embedding construction and training procedure. Other similarity measures can also be adopted. These include embeddings generated through metric learning, representations learned through neural networks, or domain-specific indicators such as demographic attributes, clinical histories, or behavioral patterns~\cite{khosla2020supervised,tao2020mining,harutyunyan2019multitask}.

\subsection{Personalized Training Objective Construction}
We construct a personalized predictive model for each target user \( u \) by integrating information from the target user’s own data and from a set of support users.
Let \( D_u \) denote the dataset of the target user, and \( \{D_j\}_{j\neq u} \) represents the datasets of all other users.
The model for user \( u\) consists of a encoder \(\phi_{\theta_u}(\cdot)\), which maps inputs into a common representation space, and a personalized prediction head \( g_{\theta_u}(\cdot) \), where \(\theta\) indicates the model parameters. The prediction for an input \( x \) is:
\begin{equation}\label{eq:wholemodel}
    \hat{y}_u = g_{\theta_u}(\phi_{\theta_u}(x))
\end{equation}

Support users are selected by the fixed initial similarity scores \( s(u,j) \). 
Each support user contributes either aligned information (when similar to the target user) or contrasting information (when dissimilar). These contributions are controlled by a set of adaptive weights \(\alpha_j\)that are learned jointly with the model parameters.

We formulate personalization as the following optimization problem:
\begin{equation}\label{eq:op}
\begin{aligned}
\min_{\theta_u, \{\alpha_j\}} \quad & \Bigg[ \lambda_p
\underbrace{L(\theta_u; D_u)}_{\text{Personal Loss}} 
+ \lambda_s \underbrace{\sum_{j \in \mathcal{S}(u)} \alpha_j s(u,j) L(\theta_u; D_j)}_{\text{Similar-User Knowledge Transfer}} \\
&+ \lambda_d \underbrace{\sum_{j \in \mathcal{D}(u)} \alpha_j [1 - s(u,j)] \, \mathcal{R}(\theta_u; D_j)}_{\text{Dissimilar-User Penalty}}
\Bigg]
\end{aligned}
\end{equation}
subject to the constraints:
\[
\sum_{j \in \mathcal{S}(u) \cup \mathcal{D}(u)} \alpha_j = 1, \quad \alpha_j \geq 0 \quad \forall j.
\]

The objective contains three components that operationalize the contributions of support users. 
The \textbf{Personal Loss} \( L(\theta_u; D_u) \) provides direct supervision from the target user’s data and ensures specificity to the user’s individual patterns.

The \textbf{Similar-User Knowledge Transfer} \( \sum_{j \in \mathcal{S}(u)} \alpha_j s(u,j) L(\theta_u; D_j) \) transfers predictive structure from users with high similarity scores. 
Data from similar users are passed through the shared encoder and the personalized head of user \(u\).
Each similar user's contribution is weighted by both a fixed similarity score \( s(u,j) \) and a dynamically adjustable weight \( \alpha_j \), which adjusts the contribution of each user based on its observed utility during training.

The \textbf{Dissimilar-User Penalty}
\( \sum_{j \in \mathcal{D}(u)} \alpha_j [1 - s(u,j)] \mathcal{R}(\theta_u; D_j) \) incorporates contrasting information from dissimilar users. The penalty \(\mathcal{R}\) is defined as:
\begin{equation}\label{eq:contrasitive}
\mathcal{R}(\theta_u; D_j) = \mathbb{E}_{x \sim D_j} \left[\max\left(0, m - \|\phi_{\theta_u}(x) - \phi_{\theta_j}(x)\|^2\right)\right],
\end{equation}
where \( m \) is a margin hyperparameter. This prevents the personal model from adopting misleading patterns while preserving useful contrastive structure. We denote by $\phi_{\theta_j}$  the encoder associated with the user $j$, and by $\phi_{\theta_u}$ the encoder of the target user $u$.

The \textbf{Adaptive Weights} \(\alpha_j\) quantify the contribution of each support user \( j \). They correct imperfections in fixed similarity scores, which may not accurately represent user relationships due to data limitations or noise, by learning user-specific relevance directly from the training objective.
The simplex constraint ensures that support-user influence is normalized.
For any data point, only the corresponding term is activated while the others are set to zero based on user type.

\subsubsection{Gradient Propagation Through Model Components}
The personalized objective in Equation~\ref{eq:op} induces structured gradients that shape the shared encoder and the personalized prediction head in complementary ways. This structure reflects the roles of personal, similar, and dissimilar support users.

\textbf{Personal Loss Propagation:}  
Gradients from  \(L(\theta_u; D_u)\) are backpropagated to both the encoder \(\phi_{\theta_u}\) and the personalized prediction head \(g_{\theta_u}\). These updates specialize the prediction head to the target user while shaping the encoder to capture user-specific structure.

\textbf{Similar-User Knowledge Transfer Loss Propagation:}  
For similar users \(j\) the loss term \(\sum_{j \in \mathcal{S}(u)} \alpha_j s(u,j) L(\theta_u; D_j)\) generates gradient that updates both target user’s encoder \(\phi_{\theta_u}(x)\) and prediction head \(g_{\theta_u}(\cdot)\).
These gradients encourage the encoder to extract shared predictive features that generalize across users with aligned patterns, and they help the personalized head map these features to accurate outputs.

\textbf{Dissimilar-User Penalty Propagation:} The penalty \(\mathcal{R}(\theta_u; D_j) \) updates only the   \(\phi_{\theta_u}\).
This preserves the specialization of the personalized head while using dissimilar users to structure the representation space. The encoder learns to separate the target user’s embeddings from those of dissimilar users, which reduces the influence of misleading correlations without interfering with the personalized prediction head.

Collectively, these gradient pathways create a coordinated optimization process. The personalized head captures the target user’s predictive characteristics, while the encoder integrates positive transfer from similar users and maintains representational separation from dissimilar users.

\subsection{Iterative Optimization}
We optimize the personalized objective in Equation~\ref{eq:op} using an alternating procedure that updates the model parameters and the adaptive support-user weights. This procedure enables the encoder and personalized head to learn from the current influence of support users, while the weights are refined based on the predictive contribution of each user.

\textbf{Step 1. Update model parameters.} Given the current adaptive weights \(\{\alpha_j\}\), we update the personalized model parameters \(\theta_u\), including both the shared encoder \(\phi_{\theta_u}\) and personalized prediction head \(g_{\theta_u}\). 
Each mini-batch is constructed by sampling data from the target user’s dataset \(D_u\) and from the datasets of support users. 
Gradients from the personal loss, similar-user transfer term, and dissimilar-user contrastive term are backpropagated into the model as described in the previous subsection. This yields parameter updates that balance user-specific supervision, aligned transfer, and contrastive regularization.

\textbf{Step 2. Update adaptive support-user weights.} With the model parameters fixed, we update the adaptive weights \(\{\alpha_j\}\) to reflect the predictive utility of each support user. For each support user \( j \), we compute the following cost:
\begin{equation}\label{eq:cost}
c_j = \lambda_s s(u,j) L(\theta_u; D_j) - \lambda_d [1 - s(u,j)] \mathcal{R}(\theta_u; D_j).
\end{equation}
where the first term quantifies how well the model predicts data from user \(j\), weighted by the \( s(u,j) \). The second term quantifies representational overlap with dissimilar users, scaled by \((1 - s(u,j))\). This cost reflects how aligned or contrasting each user is relative to the target user under the current model parameters.
The adaptive weights are then updated via softmax normalization:
\begin{equation}\label{eq:alpha}
\alpha_j = \frac{\exp(c_j)}{\sum_{k \neq u}\exp(c_k)},
\end{equation}
Users with lower cost receive higher weight, which increases their influence in the next parameter update step. 

This mechanism allows the model to identify which support users provide useful generalization signals and which provide informative contrast.
Intuitively, a lower \( c_j \) implies that user \( j \)'s data is beneficial to the target user's predictive performance (through low prediction error for similar users or a high embedding margin from dissimilar users). 




  


\section{Experiments}
Our experiments include six predictive tasks on four diverse real-world health datasets. These datasets span a wide range of healthcare personalization challenges, such as psychosocial state prediction, glycemic response estimation, and sleep forecasting. We compare the framework with existing personalization baselines using two backbone architectures.

\subsection{Datasets}

\textbf{Loneliness.}~\cite{jafarlou2024objective}  
This dataset includes multimodal data collected from individuals via mobile phones and wearable devices, with subjective loneliness measured through ecological momentary assessments (EMAs) delivered five times daily. Each participant rated their momentary loneliness on a 0–100 scale. The dataset includes features from an Oura Ring and a Samsung Galaxy Active 2 Watch (e.g., heart-rate variability, sleep stages, activity), along with smartphone-derived behavioral data (e.g., app usage, communication logs, GPS). Tasks include both regression (predicting raw scores) and binary classification (high vs. low loneliness), following prior work.

\textbf{Affect.}~\cite{labbaf2024physiological}
Collected from 20 participants over 12 months, this dataset includes daily self-reported affect scores along with multimodal data from wearables and smartphones. Features include daily sleep and activity summaries, short-window physiological recordings, and behavioral context. We evaluate both regression (positive/negative affect scores) and classification (high vs. low affect).

\textbf{CGMacro.}~\cite{gutierrez2025cgmacros}
This dataset captures short-term glucose responses from 45 adults monitored over 10 days using continuous glucose monitors and Fitbit smartwatches. Participants logged each meal with detailed macronutrient composition. The task is to predict the postprandial glucose response (iAUC) for each meal, based on meal and physiological context.

\textbf{Globem.}~\cite{xu2022globem}
A large-scale mobile sensing dataset collected from 497 individuals across four annual cohorts (2018–2021), spanning 705 person-years. Participants were passively monitored using a dedicated smartphone app and a Fitbit device over 10-week intervals. Features include daily geolocation, screen time, Bluetooth proximity, call logs, activity, and sleep. We formulate a daily forecasting task: predicting next-day sleep efficiency from the prior day's multimodal signals.

\subsection{Model Architectures and Experimental Setup}

We evaluate the framework using two encoder architectures: an MLP encoder with two hidden layers and a Transformer encoder with three attention layers. 
Both encoders are followed by a personalized prediction head with two fully connected layers and a linear output. 
We refer to these architectures as \textit{MLP} and \textit{Trans}. 
For each user, longitudinal data are split chronologically, with the first half used for training and the second half used for testing to reflect realistic deployment.

\subsection{Patient Similarity and Encoder Initialization}\label{similarity_compute}
In our experiments, we compute a simple initial patient similarity by projecting each user’s data into a latent space and then measuring pairwise distances in that space. Specifically, we train an auxiliary MLP on pooled data from all users to predict each individual's target outcomes (e.g., loneliness levels). This network maps each data point to a learned embedding, and for each user, we compute a single patient-level embedding by averaging the embeddings of their data points. Pairwise cosine similarity between these averaged embeddings is then used to define the similarity score \(s(u,j)\) for each user pair.


Additionally, we initialize the shared encoder \(\phi_{\theta_u}\) for both the \textit{MLP} and \textit{Transformer} through supervised pretraining on the aggregated training data from all users, optimized to predict target outcomes across the population. This pretraining step establishes a general-purpose embedding space that captures shared predictive patterns before any personalization occurs. After pretraining, the population-level prediction head is discarded, and the encoder is retained to initialize personalized optimization.
  
\subsection{Baseline Methods}

We compare our proposed approach against the following baseline methods to evaluate personalization performance in the health-monitoring setting.

1) \textbf{Population Model ($Pop$).} This non-personalized baseline is trained on the combined data of all users except the target user \(u\). It serves as a reference for what can be achieved without any personalization.

2) \textbf{Purely Personalized Model ($Per_{pure}$):}  
An N-of-1 baseline trained solely on the target individual's own historical data, explicitly ignoring other patients’ data~\cite{nan2024personalized}. This method illustrates personalization performance limited to individual data availability.

3) \textbf{Merged Personalized Model ($Per_{merge}$):}  
Personalized models are created by naively combining the target individual's data with the entire population dataset. This baseline treats all data uniformly, without differentiating between personal and population data.

4) \textbf{Transfer-based Personalized Model ($Per_{trans}$).}  
This baseline follows a standard sequential transfer-learning scheme. 
A population model is pretrained on data from all users except the target user \(u\), and the personal head is then fine-tuned using \(u\)’s data~\cite{oh2021fedbabu}. 
All external users contribute equally during pretraining, and no user selection is performed.

5) \textbf{Cluster-based Personalized Model ($Per_{c}$).}  
This baseline restricts personalization to users within the same cluster as the target user. 
Users are embedded by averaging their training feature vectors and grouped using $k$-means clustering with \(K=5\). 
For a target user \(u\), the model is trained using data pooled from users in the same cluster, reflecting a common strategy to reduce negative transfer by using only similar users~\cite{yue2014personalized}.

6) \textbf{Similarity-Weighted Personalized Model ($Per_{we}$).}  
This baseline weights each external user’s contribution using a fixed similarity score relative to the target user~\cite{welch2022leveraging}. 
The model is trained on all users’ data, with each user’s loss scaled by a precomputed similarity score \(s(u,j)\). 
The similarity scores remain fixed throughout training and do not adapt to model learning.

\section{Results}
We report the performance using root mean squared error (RMSE) for the regression task and accuracy (Acc) for the classification task. We assess data efficiency by examining performance as a function of personal training data availability. Moreover, we analyze adaptive similarity weights (\(\alpha_j\)) to demonstrate interpretability and practical utility in guiding future data collection and interventions.

\subsection{Overall Results}

\begin{table*}[htbp]
\centering
\begin{tabular}{llc|ccccccc}
\toprule
Dataset & Task & Encoder &\textbf{Ours} & $Pop$ & $Per_{pure}$ & $Per_{merge}$ & $Per_{trans}$ & $Per_{c}$ & $Per_{we}$ \\
\midrule
\multirow{4}{*}{Loneliness} 
& \multirow{2}{*}{Reg (RMSE$\downarrow$)}  
&\textit{MLP} 
& \textbf{13.85} & 33.20 & 18.81 & 21.75 & 17.77 & 17.10 & 16.85 \\
& &\textit{Trans} 
& \textbf{15.13} & 42.10 & 22.70 & 25.60 & 20.33 & 19.90 & 19.60 \\
 \cmidrule{2-10}
& \multirow{2}{*}{Clas (Acc$\uparrow$)} 
&\textit{MLP} 
& \textbf{0.806} & 0.676 & 0.782 & 0.788 & 0.794 & 0.796 & 0.799 \\
& &\textit{Trans} 
& \textbf{0.802} & 0.780 & 0.760 & 0.770 & 0.769 & 0.772 & 0.775 \\
\midrule
\multirow{8}{*}{Affect} 
& \multirow{2}{*}{PA Reg (RMSE$\downarrow$)} 
&\textit{MLP}  
& \textbf{15.56} & 27.08 & 17.16 & 22.47 & 18.42 & 18.05 & 17.85 \\
& &\textit{Trans} 
& 23.10 & 21.67 & 22.94 & 19.41 & \textbf{13.60} & 15.10 & 14.40 \\
  \cmidrule{2-10}
& \multirow{2}{*}{NA Reg (RMSE$\downarrow$)}  
&\textit{MLP} 
& \textbf{15.45} & 23.05 & 17.16 & 19.59 & 18.20 & 17.95 & 17.70 \\
& &\textit{Trans} 
& 26.25 & 19.33 & \textbf{16.41} & 18.22 & 24.86 & 21.40 & 20.80 \\
   \cmidrule{2-10}
&  \multirow{2}{*}{PA Clas (Acc$\uparrow$)} 
&\textit{MLP} 
& \textbf{0.826} & 0.529 & 0.812 & 0.544 & 0.817 & 0.808 & 0.812 \\
&  &\textit{Trans} 
& \textbf{0.817} & 0.482 & 0.745 & 0.762 & 0.762 & 0.758 & 0.761 \\
   \cmidrule{2-10}
& \multirow{2}{*}{NA Clas (Acc$\uparrow$)} 
&\textit{MLP} 
& \textbf{0.778} & 0.530 & 0.756 & 0.541 & 0.691 & 0.728 & 0.735 \\
&  &\textit{Trans} 
& \textbf{0.756} & 0.541 & 0.698 & 0.681 & 0.688 & 0.690 & 0.695 \\
\midrule
\multirow{2}{*}{CGMacro} 
& \multirow{2}{*}{Reg (RMSE$\downarrow$)}  
&\textit{MLP} 
& 5825.1 & 2760.2 & 3825.1 & \textbf{2538.4} & 2988.2 & 2755.0 & 2905.0 \\
&  &\textit{Trans} 
& 2523.7 & 2824.4 & 5011.3 & \textbf{2486.4} & 2822.5 & 2705.0 & 2795.0 \\
\midrule
\multirow{2}{*}{Globem}  
& \multirow{2}{*}{RReg (RMSE$\downarrow$)} 
&\textit{MLP}  
& \textbf{0.340} & 0.399 & 0.451 & 0.387 & 0.390 & 0.366 & 0.372 \\
&  &\textit{Trans} 
& \textbf{0.342} & 0.433 & 0.426 & 0.376 & 0.387 & 0.362 & 0.369 \\
\bottomrule
\end{tabular}

\caption{Performance comparison across datasets and tasks. For Loneliness and Affect datasets, "Reg" denotes RMSE (lower is better) and "Clas" indicates accuracy (higher is better). PA and NA refer to positive affect and negative affect, respectively. CGMacro and Globem show regression RMSE values (lower is better). Best performances are highlighted in bold.}
\label{tab:overallresults}
\end{table*}

Table~\ref{tab:overallresults} summarizes the performance across four diverse digital health datasets and multiple prediction tasks. 
Overall, the proposed framework outperforms non-personalized and traditional personalized baseline methods on the majority of evaluated tasks.
On both the \textbf{Loneliness} and \textbf{Affect} datasets, our method achieves performance improvements over baseline methods using the same model architectures. 
For the Loneliness dataset, it narrows the performance gap previously observed between deep learning models and traditional machine learning methods~\cite{yang2023loneliness}.
Similarly, on the Affect dataset, the proposed method reduces prediction error for both positive and negative affect tasks compared to other personalization methods. 
We further observe that similarity-restricted baselines, including the cluster-based model ($Per_c$) and the fixed similarity-weighted model ($Per_{we}$), generally outperform naive population merging but remain consistently inferior to the proposed approach. This suggests that static similarity constraints or hard clustering can partially mitigate negative transfer, but lack the flexibility to adaptively modulate user influence or exploit informative contrast from dissimilar users during training.

We observed that dataset characteristics can heavily influence outcomes, as illustrated by the \textbf{CGMacro} dataset.
Performance on CGMacro was notably weaker across all tested methods.
The regression RMSE was significantly higher compared to other datasets, reflecting fundamental challenges due to limited data points per patient (approximately 10 days of data) and inherent data variability. Personalization approaches struggled to achieve stable and effective models under these highly constrained conditions. However, we observed that the learned adaptive similarity weights (\(\alpha_j\)) exhibited high variability and fluctuated throughout training. This contrasts with the stable final weights observed in richer datasets.
Such behavior suggests a unique practical utility of the adaptive mechanism in explicitly signaling insufficient or unreliable training conditions.

In contrast, the larger-scale \textbf{Globem} dataset highlights the advantages of the proposed method in data-rich conditions where it manifests best performance.
This indicates the proposed method's effectiveness in leveraging extensive and diverse patient populations to robustly inform personalized models.
The rich patient similarity structure derived from abundant longitudinal data points per patient and a large participant pool (over 140 individuals) enables our adaptive optimization approach with more accurate patient similarities.

In addition, we find that simpler \textit{MLP} outperform \textit{Trans} on most tasks, particularly in small-data regimes such as Loneliness and Affect prediction. This aligns with prior findings that traditional machine learning models can outperform deep models when data is limited~\cite{yang2023loneliness,jafarlou2022objective}. As data availability increases, as in the \textbf{Globem} dataset, this performance gap narrows under our method, suggesting that more complex architectures may benefit from richer data contexts.

\subsection{Ablation Study}

\begin{table}[tbp]
\centering
\small
\begin{tabular}{llcccc}
\toprule
\textbf{Dataset} & \textbf{Task} & \textbf{Full} & \textbf{w/o $\mathcal{R}$} & \textbf{w/o $\mathcal{L}$} & \textbf{w/o $\alpha$} \\
\midrule
\multirow{2}{*}{Loneliness} & Reg & \textbf{13.85} & 20.31 & 19.05 & 16.92 \\
                            & Clas    & \textbf{0.806} & 0.799 & 0.778 & 0.801 \\
\midrule
\multirow{4}{*}{Affect} & PA Reg & \textbf{15.56} & 19.70 & 22.46 & 15.99 \\
                        & NA Reg  & \textbf{15.45} & 17.43 & 23.11 & 16.65 \\
                        & PA Clas  & \textbf{0.826} & 0.783 & 0.787 & 0.775 \\
                        & NA Clas   & \textbf{0.778} & 0.747 & 0.718 & 0.752 \\
\midrule
CGMacro & Reg & 5825.1 & 5217.1 & 5901.3 & \textbf{5106.7} \\
\midrule
Globem  & Reg& \textbf{0.340} & 0.383 & 0.391 & 0.343 \\
\bottomrule
\end{tabular}

\caption{
Ablation study comparing performance when key information is removed from training the \textit{MLP} model “$\mathcal{R}$” removes the penalty term from dissimilar users; “$\mathcal{L}$” removes knowledge transfer from similar users; “w/o $\alpha$” uses fixed similarity weights. Bold values indicate best performance.
}
\label{tab:ablation}
\end{table}

\begin{figure*}[htb]
    \centering
    \includegraphics[width=0.9\linewidth]{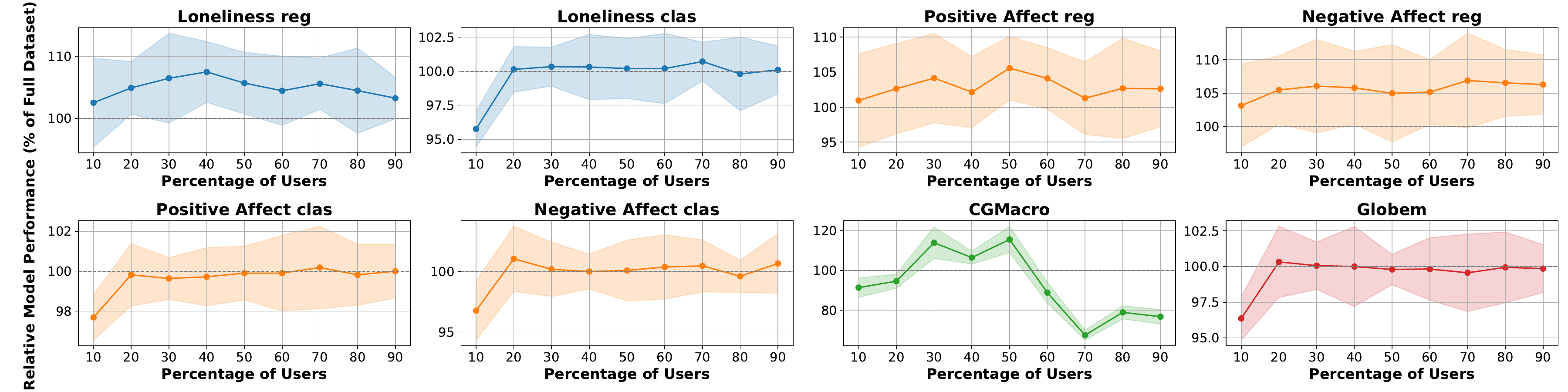}
    \caption{Relative model performance across tasks when varying the percentage of users used for training.}
    \label{fig:dataeff}
\end{figure*}
On the \textit{MLP} structure, we compare the full framework against three ablated variants: (1) removing the dissimilar-user penalty term \textbf{(w/o $\mathcal{R}$)}, (2) removing the similar-user knowledge transfer term (\textbf{(w/o $\mathcal{L}$)}, and (3) removing the adaptive weighting mechanism (i.e., using fixed similarity scores without updating $\alpha_j$ during training, \textbf{(w/o $\alpha$)}. The comparisons are presented in Table~\ref{tab:ablation}.

Both the similar-user knowledge transfer and the dissimilar-user penalty terms contribute meaningfully to performance, though their relative importance varies across tasks and datasets. 
In several cases (e.g., Affect regression and Globem), removing either term leads to significant degradation, which supports our hypothesis that effective personalization requires leveraging both shared patterns from similar individuals and contrastive signals from dissimilar individuals. 

The adaptive weighting mechanism further improves performance on most datasets, particularly those where initial similarity metrics are less informative or where the population is more heterogeneous (e.g., Affect).
This suggests that static similarity measures alone may not capture nuanced relationships relevant for personalization and that adaptively learning inter-user weights during training is critical for adjusting contributions in context-sensitive ways.

\subsection{Data Efficiency Analysis}

We analyze how the amount of population data used during training affects personalization performance. 
For each target patient, we vary the threshold of selecting support users included in training by retaining only the most similar and most dissimilar patients based on initial similarity, while excluding the middle range.
For example, using 20\% includes only the top 10\% most similar and top 10\% most dissimilar patients.

As shown in Figure~\ref{fig:dataeff}, models trained on selected subsets often outperform those trained on the full population. This trend appears consistently across tasks and datasets. 
These results support the assumption behind support users that the most relevant information for personalization comes from the support users who are either closely aligned or meaningfully different.

\subsection{Adaptive Weights Interpretation}

\begin{figure}[tb]
    \centering
    \begin{subfigure}[t]{0.475\linewidth}
        \includegraphics[width=\linewidth]{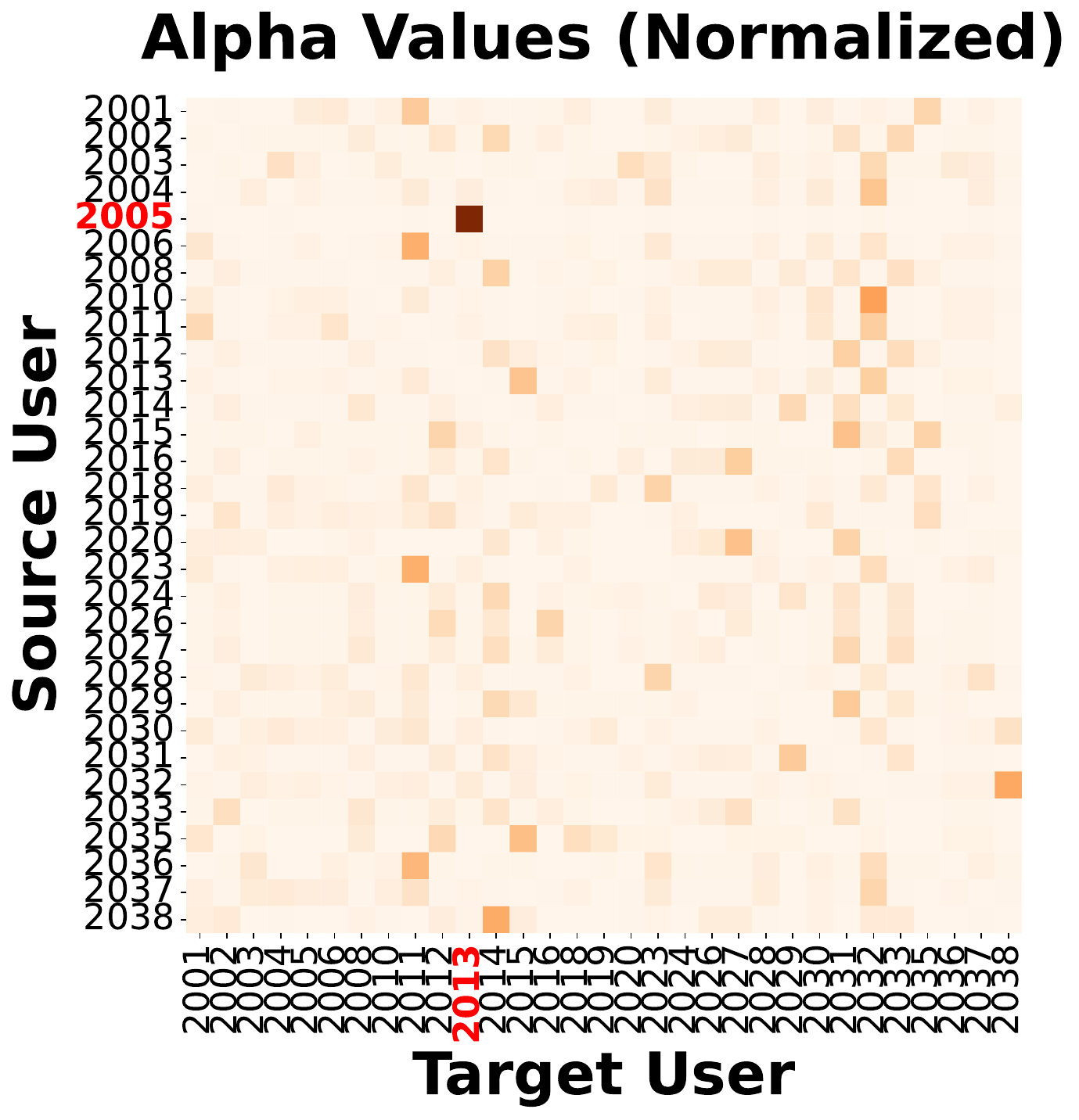}
        \caption{Final adaptive weights (\(\alpha_j\))}
        \label{fig:alpha_matrix}
    \end{subfigure}
    \hfill
    \begin{subfigure}[t]{0.5\linewidth}
        \includegraphics[width=\linewidth]{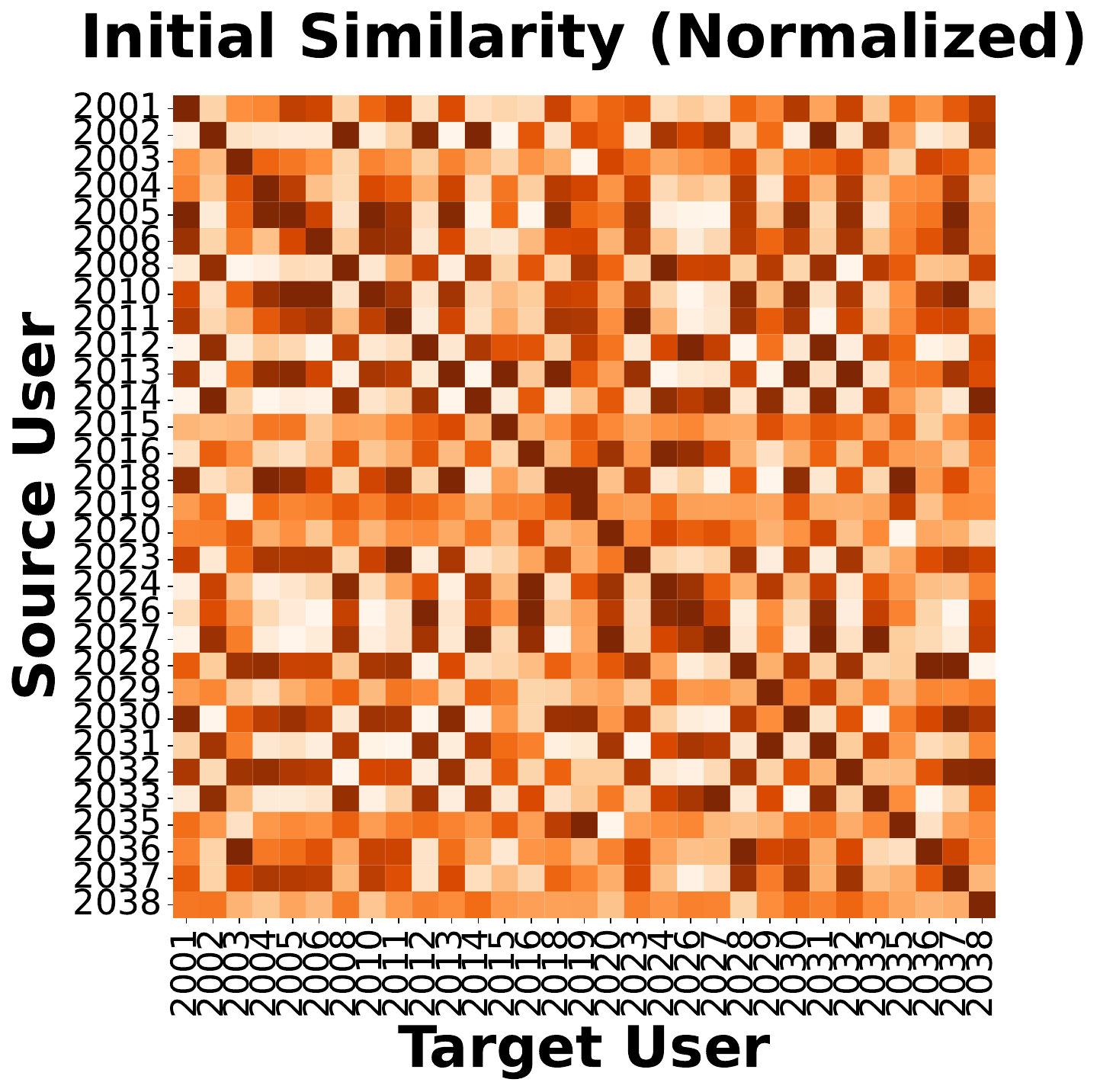}
        \caption{Initial patient similarity}
        \label{fig:similarity_matrix}
    \end{subfigure}
    \caption{
Comparison between learned weights and initial similarity scores, normalized for visualization. Each row represents a source user contributing to the target user's model (columns).
    }
    \label{fig:alphas}
\end{figure}

We examine the learned (\(\alpha_j\)) between support users at the end of training. Figure~\ref{fig:alphas} shows heatmaps of the initial similarity scores and the final \(\alpha_j\) values for each target individual across a cohort of patients.
Each column represents a target user's personal model, each row corresponds to a source user, and the color of each cell indicates the normalized (\(\alpha_j\)) value after training is complete.

We observe that the initial similarity matrix is relatively dense, the learned \(\alpha_j\) matrix is much sparser and readable as it selectively concentrates influence on a small subset of support users.
For instance, user 2005 assigns high \(\alpha_j\) values to just a few other target patients, especially to user 2013's.

The learned \(\alpha_j\) values could carry practical significance. 
A high \(\alpha_j\) for a particular individual \(j\) indicates that their data contributes substantially to the training of the target user \(u\)'s model. 
In other words, for further improving the personal model for user \(u\), it would be beneficial to prioritize collecting more data not only from \(u\) but also from those patients with high \(\alpha_j\), as their data has been demonstrated to be particularly useful during training.
On the other hand, by summing the \(\alpha_j\) values that a given patient receives across all models, we can estimate how broadly valuable that individual’s data is to others. Users with consistently high total \(\alpha_j\) across models are influential to many, suggesting they would carry more representative information for personal models at the current stage. 
This insight can guide data collection strategies.


\section{Conclusion}

We proposed a personalization framework that adaptively integrates individual data with selectively weighted population data for digital health modeling. 
Across six tasks on four real-world datasets, the framework consistently outperformed population-level, transfer-based, and similarity-based baselines, achieving up to 10\% lower RMSE on large-scale data and approximately 25\% lower RMSE in low-data settings. 
We found that simpler MLP encoders are more effective under limited data, while architectural differences diminish as data availability increases. 
The learned adaptive weights improve data efficiency and provide interpretable guidance for targeted data selection and personalized model development.

\bibliographystyle{named}
\bibliography{custom}

@article{yang2024chatdiet,
  title={ChatDiet: Empowering personalized nutrition-oriented food recommender chatbots through an LLM-augmented framework},
  author={Yang, Zhongqi and Khatibi, Elahe and Nagesh, Nitish and Abbasian, Mahyar and Azimi, Iman and Jain, Ramesh and Rahmani, Amir M},
  journal={Smart Health},
  volume={32},
  pages={100465},
  year={2024},
  publisher={Elsevier}
}

@article{nan2024personalized,
  title={Personalized machine learning-based prediction of wellbeing and empathy in healthcare professionals},
  author={Nan, Jason and Herbert, Matthew S and Purpura, Suzanna and Henneken, Andrea N and Ramanathan, Dhakshin and Mishra, Jyoti},
  journal={Sensors},
  volume={24},
  number={8},
  pages={2640},
  year={2024},
  publisher={MDPI}
}

@article{harutyunyan2019multitask,
  title={Multitask learning and benchmarking with clinical time series data},
  author={Harutyunyan, Hrayr and Khachatrian, Hrant and Kale, David C and Ver Steeg, Greg and Galstyan, Aram},
  journal={Scientific data},
  volume={6},
  number={1},
  pages={96},
  year={2019},
  publisher={Nature Publishing Group UK London}
}

@article{tao2020mining,
  title={Mining health knowledge graph for health risk prediction},
  author={Tao, Xiaohui and Pham, Thuan and Zhang, Ji and Yong, Jianming and Goh, Wee Pheng and Zhang, Wenping and Cai, Yi},
  journal={World Wide Web},
  volume={23},
  number={4},
  pages={2341--2362},
  year={2020},
  publisher={Springer}
}

@article{khosla2020supervised,
  title={Supervised contrastive learning},
  author={Khosla, Prannay and Teterwak, Piotr and Wang, Chen and Sarna, Aaron and Tian, Yonglong and Isola, Phillip and Maschinot, Aaron and Liu, Ce and Krishnan, Dilip},
  journal={Advances in neural information processing systems},
  volume={33},
  pages={18661--18673},
  year={2020}
}

@inproceedings{tazarv2021personalized,
  title={Personalized stress monitoring using wearable sensors in everyday settings},
  author={Tazarv, Ali and Labbaf, Sina and Reich, Stephanie M and Dutt, Nikil and Rahmani, Amir M and Levorato, Marco},
  booktitle={2021 43rd Annual International Conference of the IEEE Engineering in Medicine \& Biology Society (EMBC)},
  pages={7332--7335},
  year={2021},
  organization={IEEE}
}

@inproceedings{wu2023personalized,
  title={Personalized federated learning with parameter propagation},
  author={Wu, Jun and Bao, Wenxuan and Ainsworth, Elizabeth and He, Jingrui},
  booktitle={Proceedings of the 29th ACM SIGKDD conference on knowledge discovery and data mining},
  pages={2594--2605},
  year={2023}
}

@inproceedings{hahn2022connecting,
  title={Connecting low-loss subspace for personalized federated learning},
  author={Hahn, Seok-Ju and Jeong, Minwoo and Lee, Junghye},
  booktitle={Proceedings of the 28th ACM SIGKDD Conference on Knowledge Discovery and Data Mining},
  pages={505--515},
  year={2022}
}

@inproceedings{zhang2024gpfedrec,
  title={Gpfedrec: Graph-guided personalization for federated recommendation},
  author={Zhang, Chunxu and Long, Guodong and Zhou, Tianyi and Zhang, Zijian and Yan, Peng and Yang, Bo},
  booktitle={Proceedings of the 30th ACM SIGKDD Conference on Knowledge Discovery and Data Mining},
  pages={4131--4142},
  year={2024}
}

@article{kolluri2016revolutionizing,
  title={Revolutionizing Healthcare With AI: Personalized Medicine: Predictive},
  author={Kolluri, VENKATESWARANAIDU},
  journal={JETIR-Int. J. Emerg. Technol. Innov. Res},
  volume={3},
  number={11},
  pages={2349--5162},
  year={2016}
}

@article{deng2020adaptive,
  title={Adaptive personalized federated learning},
  author={Deng, Yuyang and Kamani, Mohammad Mahdi and Mahdavi, Mehrdad},
  journal={arXiv preprint arXiv:2003.13461},
  year={2020}
}

@article{wu2024diversity,
  title={The diversity bonus: Learning from dissimilar distributed clients in personalized federated learning},
  author={Wu, Xinghao and Liu, Xuefeng and Niu, Jianwei and Zhu, Guogang and Tang, Shaojie and Li, Xiaotian and Cao, Jiannong},
  journal={arXiv preprint arXiv:2407.15464},
  year={2024}
}

@inproceedings{chen2024personalized,
  title={Personalized federated learning with attention-based client selection},
  author={Chen, Zihan and Li, Jundong and Shen, Cong},
  booktitle={ICASSP 2024-2024 IEEE International Conference on Acoustics, Speech and Signal Processing (ICASSP)},
  pages={6930--6934},
  year={2024},
  organization={IEEE}
}

@inproceedings{huang2021personalized,
  title={Personalized cross-silo federated learning on non-iid data},
  author={Huang, Yutao and Chu, Lingyang and Zhou, Zirui and Wang, Lanjun and Liu, Jiangchuan and Pei, Jian and Zhang, Yong},
  booktitle={Proceedings of the AAAI conference on artificial intelligence},
  volume={35},
  number={9},
  pages={7865--7873},
  year={2021}
}

@article{zhang2020personalized,
  title={Personalized federated learning with first order model optimization},
  author={Zhang, Michael and Sapra, Karan and Fidler, Sanja and Yeung, Serena and Alvarez, Jose M},
  journal={arXiv preprint arXiv:2012.08565},
  year={2020}
}

@article{yi2024pfedmoe,
  title={pFedMoE: Data-level personalization with mixture of experts for model-heterogeneous personalized federated learning},
  author={Yi, Liping and Yu, Han and Ren, Chao and Zhang, Heng and Wang, Gang and Liu, Xiaoguang and Li, Xiaoxiao},
  journal={arXiv preprint arXiv:2402.01350},
  year={2024}
}

@inproceedings{liu2023auxo,
  title={Auxo: Efficient federated learning via scalable client clustering},
  author={Liu, Jiachen and Lai, Fan and Dai, Yinwei and Akella, Aditya and Madhyastha, Harsha V and Chowdhury, Mosharaf},
  booktitle={Proceedings of the 2023 ACM Symposium on Cloud Computing},
  pages={125--141},
  year={2023}
}

@article{chen2018federated,
  title={Federated meta-learning with fast convergence and efficient communication},
  author={Chen, Fei and Luo, Mi and Dong, Zhenhua and Li, Zhenguo and He, Xiuqiang},
  journal={arXiv preprint arXiv:1802.07876},
  year={2018}
}

@inproceedings{welch2022leveraging,
  title={Leveraging similar users for personalized language modeling with limited data},
  author={Welch, Charles and Gu, Chenxi and Kummerfeld, Jonathan K and Perez-Rosas, Veronica and Mihalcea, Rada},
  booktitle={Proceedings of the 60th Annual Meeting of the Association for Computational Linguistics (Volume 1: Long Papers)},
  pages={1742--1752},
  year={2022}
}

@article{t2020personalized,
  title={Personalized federated learning with moreau envelopes},
  author={T Dinh, Canh and Tran, Nguyen and Nguyen, Josh},
  journal={Advances in neural information processing systems},
  volume={33},
  pages={21394--21405},
  year={2020}
}

@inproceedings{li2021ditto,
  title={Ditto: Fair and robust federated learning through personalization},
  author={Li, Tian and Hu, Shengyuan and Beirami, Ahmad and Smith, Virginia},
  booktitle={International conference on machine learning},
  pages={6357--6368},
  year={2021},
  organization={PMLR}
}

@article{li2021fedbn,
  title={Fedbn: Federated learning on non-iid features via local batch normalization},
  author={Li, Xiaoxiao and Jiang, Meirui and Zhang, Xiaofei and Kamp, Michael and Dou, Qi},
  journal={arXiv preprint arXiv:2102.07623},
  year={2021}
}

@article{oh2021fedbabu,
  title={Fedbabu: Towards enhanced representation for federated image classification},
  author={Oh, Jaehoon and Kim, Sangmook and Yun, Se-Young},
  journal={arXiv preprint arXiv:2106.06042},
  year={2021}
}

@inproceedings{yue2014personalized,
  title={Personalized collaborative clustering},
  author={Yue, Yisong and Wang, Chong and El-Arini, Khalid and Guestrin, Carlos},
  booktitle={Proceedings of the 23rd international conference on World wide web},
  pages={75--84},
  year={2014}
}

@article{jain2021differentially,
  title={Differentially private model personalization},
  author={Jain, Prateek and Rush, John and Smith, Adam and Song, Shuang and Guha Thakurta, Abhradeep},
  journal={Advances in neural information processing systems},
  volume={34},
  pages={29723--29735},
  year={2021}
}

@article{lee2023fedl2p,
  title={Fedl2p: Federated learning to personalize},
  author={Lee, Royson and Kim, Minyoung and Li, Da and Qiu, Xinchi and Hospedales, Timothy and Husz{\'a}r, Ferenc and Lane, Nicholas},
  journal={Advances in Neural Information Processing Systems},
  volume={36},
  pages={14818--14836},
  year={2023}
}

@article{jafarlou2022objective,
  title={Objective Prediction of Tomorrow's Affect Using Multi-Modal Physiological Data and Personal Chronicles: A Study of Monitoring College Student Well-being in 2020},
  author={Jafarlou, Salar and Lai, Jocelyn and Mousavi, Zahra and Labbaf, Sina and Jain, Ramesh and Dutt, Nikil and Borelli, Jessica and Rahmani, Amir},
  journal={arXiv preprint arXiv:2201.11230},
  year={2022}
}

@article{labbaf2024physiological,
  title={Physiological and emotional assessment of college students using wearable and mobile devices during the 2020 COVID-19 lockdown: an intensive, longitudinal dataset},
  author={Labbaf, Sina and Abbasian, Mahyar and Nguyen, Brenda and Lucero, Matthew and Ahmed, Maryam Sabah and Yunusova, Asal and Rivera, Alexander and Jain, Ramesh and Borelli, Jessica L and Dutt, Nikil and others},
  journal={Data in Brief},
  volume={54},
  pages={110228},
  year={2024},
  publisher={Elsevier}
}

@article{gutierrez2025cgmacros,
  title={CGMacros: a scientific dataset for personalized nutrition and diet monitoring},
  author={Gutierrez-Osuna, Ricardo and Kerr, David and Mortazavi, Bobak and Das, Anurag},
  journal={Scientific Data (under review)},
  year={2025}
}

@article{xu2022globem,
  title={GLOBEM dataset: multi-year datasets for longitudinal human behavior modeling generalization},
  author={Xu, Xuhai and Zhang, Han and Sefidgar, Yasaman and Ren, Yiyi and Liu, Xin and Seo, Woosuk and Brown, Jennifer and Kuehn, Kevin and Merrill, Mike and Nurius, Paula and others},
  journal={Advances in neural information processing systems},
  volume={35},
  pages={24655--24692},
  year={2022}
}

@article{fallah2020personalized,
  title={Personalized federated learning: A meta-learning approach},
  author={Fallah, Alireza and Mokhtari, Aryan and Ozdaglar, Asuman},
  journal={arXiv preprint arXiv:2002.07948},
  year={2020}
}

@inproceedings{mcmahan2017communication,
  title={Communication-efficient learning of deep networks from decentralized data},
  author={McMahan, Brendan and Moore, Eider and Ramage, Daniel and Hampson, Seth and y Arcas, Blaise Aguera},
  booktitle={Artificial intelligence and statistics},
  pages={1273--1282},
  year={2017},
  organization={PMLR}
}

@article{siirtola2019incremental,
  title={Incremental learning to personalize human activity recognition models: the importance of human AI collaboration},
  author={Siirtola, Pekka and R{\"o}ning, Juha},
  journal={Sensors},
  volume={19},
  number={23},
  pages={5151},
  year={2019},
  publisher={MDPI}
}

@article{jafarlou2024objective,
  title={Objective monitoring of loneliness levels using smart devices: A multi-device approach for mental health applications},
  author={Jafarlou, Salar and Azimi, Iman and Lai, Jocelyn and Wang, Yuning and Labbaf, Sina and Nguyen, Brenda and Qureshi, Hana and Marcotullio, Christopher and Borelli, Jessica L and Dutt, Nikil D and others},
  journal={Plos one},
  volume={19},
  number={6},
  pages={e0298949},
  year={2024},
  publisher={Public Library of Science San Francisco, CA USA}
}

@inproceedings{yang2023loneliness,
  title={Loneliness forecasting using multi-modal wearable and mobile sensing in everyday settings},
  author={Yang, Zhongqi and Azimi, Iman and Jafarlou, Salar and Labbaf, Sina and Borelli, Jessica and Dutt, Nikil and Rahmani, Amir M},
  booktitle={2023 IEEE 19th International Conference on Body Sensor Networks (BSN)},
  pages={1--4},
  year={2023},
  organization={IEEE}
}

@article{yang2024integrating,
  title={Integrating wearable sensor data and self-reported diaries for personalized affect forecasting},
  author={Yang, Zhongqi and Wang, Yuning and Yamashita, Ken S and Khatibi, Elahe and Azimi, Iman and Dutt, Nikil and Borelli, Jessica L and Rahmani, Amir M},
  journal={Smart Health},
  volume={32},
  pages={100464},
  year={2024},
  publisher={Elsevier}
}

@article{yang2025personalized,
  title={Personalized Causal Graph Reasoning for LLMs: A Case Study on Dietary Recommendations},
  author={Yang, Zhongqi and Rahmani, Amir},
  journal={arXiv preprint arXiv:2503.00134},
  year={2025}
}

@article{johnson2021precision,
  title={Precision medicine, AI, and the future of personalized health care},
  author={Johnson, Kevin B and Wei, Wei-Qi and Weeraratne, Dilhan and Frisse, Mark E and Misulis, Karl and Rhee, Kyu and Zhao, Juan and Snowdon, Jane L},
  journal={Clinical and translational science},
  volume={14},
  number={1},
  pages={86--93},
  year={2021},
  publisher={Wiley Online Library}
}

\end{document}